# Extracting and Validating Explanatory Word Archipelagoes using Dual Entropy

Yukio Ohsawa

Department of Systems Innovation, School of Eng. The University of Tokyo
Ohsawa -at- sys.t.u-tokyo.ac.jp

## Abstract

The logical connectivity of text is represented by the connectivity of words that form *archipelagoes*. Here, each archipelago is a sequence of islands of the occurrences of a certain word. An island here means the local sequence of sentences where the word is emphasized, and an archipelago of a length comparable to the target text is extracted using the co-variation of entropy A (*the window-based entropy*) on the distribution of the word's occurrences with the width of each time window. Then, the logical connectivity of text is evaluated on entropy B (*the graph-based entropy*) computed on the distribution of sentences to connected word-clusters obtained on the co-occurrence of words. The results show the parts of the target text with words forming *archipelagoes* extracted on *entropy A*, without learned or prepared knowledge, form an explanatory part of the text that is of smaller entropy B than the parts extracted by the baseline methods.

## 1 Introduction

The coherent flow of context in a text has been regarded as a feature of readability of the text. The contextual coherence has been evaluated using machine learning techniques. Basically, the techniques evaluate the continuity of the learned word meanings in the text, mixing the sentence-to-sentence interactions in the target text in recent developments. On the other hand, the aim of this paper is not to evaluate or improve the readability meaning the easiness to read the document but to extract logically explanatory parts of the target document. A word that enables us to explain the text is called an *explanatory word* here. It should be also noted that *explanatory* means a different concept from *explainable* that is a focus of studies and applications of machine learning [Baehrens et al 2010, Gunning and Aha 2019]. When one says X is *explainable*, it means *X* can be explained. On the other hand, when one says X is *explanatory*, it means *X* is useful information for explaining something, as in the usage *explanatory hypothesis* or *explanatory coherence* [Thagard 1989, Ng and Mooney 1990] for explaining the causalities of observed events and *explanatory sign* [Ohsawa 2019] that means an event on which the latent dynamics can be explained.

From the viewpoint of text analysis, the two concepts, i.e. readable due to coherence and explanatory due to logical flow, differ in that the former means there are less discontinuous contextual shifts that may cause overload on the readers' reasoning whereas the latter allows discontinuous occurrence of new scenes and contexts with the progress of text. For example, "*I have books about mathematics. The books are hard to understand. So I also read other books of when I am tired., ...*" is easy to read because the three sentences are covered by the context of books. Thus, the context coherence has been evaluated on the similarity of the topics of the parts of text in topic models [Bleiand Lafferty 2006, Iwata et al 2009], and of embedding vectors of consecutive words or sentences applying the deep learning approaches such as CNN[Cui et al 2017], RNN/LSTM [Strube, Mohsen Mesgar 2018], and attention-based model for translation maximizing the contextual coherence [Xiong et al 2018]. On the other hand, "*I have a book about mathematics. The concepts in discrete mathematics are hard to understand. One of the concepts is graph theory., ...*" is less easy but the concepts are logically connected. That is, the first sentence shows a concept "book of mathematics", and the second presents "*discrete*" and "*concept"* related to the first concept. Then, the third explains "*graph theory*" as a "*concept.*"

The difference between the two examples above is that the latter gives a new concept sentence by sentence, whereas the former stays in the concept "books." A document for creative work should be the one like the latter, that is our interest here. In such a document, context of neighboring sentences may differ and the document may be hard to read. However, the document may have a logical structure to derive a new theory. In this paper, we propose an approach to extracting explanatory information rather than words for highlighting coherent sequences in or for classification of the target text.

Here we summarize the idea of word archipelagoes in Section 2. The expected effect of words that form archipelagoes as explanatory information about the document is illustrated. In Section 3, the computation method of window-based entropy (*entropy A*) for extracting word archipelagoes is proposed as well as the method to use graph-based entropy (*entropy B*) for evaluating the explanatoriness of the extracted part of the text. Let us position the technical aspect of this paper relative to the previous work for information extraction from the text in Section 2 and in the experimental

Section 4. We do not go ahead to the comparison with summarization technologies because our output, for the time being, is keyword-set although this is not our final goal. In Section 4, the performance of the presented method is shown to outperform keyword unsupervised extraction methods.

## 2 Extraction of word archipelagoes

### 2.1 We do not rank, connect words

The machine-learning approaches are capable to consider contexts under and relationships between different features to make predictions of forthcoming words or latent topics. Support vector machines, conditional random fields, and variations of deep learning have been employed for learning complex patterns from a text with weighting words or features in a sequence of words in the text [Wu 2007, Tang et al 2019]. An unsupervised method for key-phrase extraction, leveraging sentence embeddings, is shown to increase the coverage and the diversity of selected key-phrases from a large set of data [Bennani-Smires, K., Musat, C. 2018].

However, our problem to extract explanatory information can be hardly defined as a problem for machine learning at least for two reasons. First, our final goal is not ranking, labeling, or classification of words or a sequence of words but to extract the logical flow in the text. That is, we cannot predefine the classes or ranking to learn because there is no concept "class" or "correctness" we can define from the position of supervisors. In this sense, we neither aim to rank words which means to compare the correctness or importance of words (meaning large islands, high mountains, or central hub cities in the maps of words) but to find connective words which means to find essential bridges between islands to form a paths of logical structure of the entire text. Words that bridge between highly ranked words may be ranked lower but are the core target of this study. Second, and similarly, if we consider the unsupervised learning approach, we should clarify clustering is not the way for us because our aim is to show out the connectors of islands rather than islands themselves as clusters. In this sense, we share the vision with [Campos et al 2020]. For example, given a set of novels of Maurice LeBranc, one about the appearance of the genius thief Arsene Lupin and the other about Lupin's being arrested by detective Ganimard can be combined to believe Ganimard's super-high level talent. This belief is more interesting than just classifying novels into about successes and failures of Lupin for one interested in the talented persons in Lupin's series. Furthermore, LeBranc's next novel about Lupin's jailbreak can be classified as Lupin's successes. However, adding just one more success of Lupin's is not as interesting as a revenge to his strong enemy Ganimard.

Methods for keyword extraction in Graph-based Approaches can be one way to obtain explanatory information. Here, a text is modeled as a graph where words are regarded as vertices connected by directed or no-directed labeled edges, on the idea to measure the importance of vertices, corresponding to words' importance, taking into consideration the structure of the graph. This structure of the graph can be regarded as an approximation of the logical structure, so graph-based keyword extraction [Matsuo et al 2001, Beliga et al 2015] has been established. Graph-based is extended to approaches to extract key-phrases related to the major topics within a text [Yinga and Qingping 2017]. Furthermore, an extensive approach to explore keywords using statistical methods and wordnet-based pattern evaluation has been presented [Dostal and Jezek 2010]. The RAKE algorithm considers the degree and the count of words in the co-occurrence structure, so can be positioned as graph-based if we regard the co-occurrence structure as a graph [Rose et al 2010]. TextRank, an algorithm borrowing the basic idea from PageRank, has been used in keyword extraction and text summarization [Mihalcea and Tarau 2004].

### 2.1 A word that forms an archipelago

If the exemplified texts in the introduction last for a large number (tens or hundreds) of sentences, the concept "graph theory" will turn out to play just a transient role in the entire text as far as it does not appear later again. The word or the phrase tends to contribute to the mainstream of the text if (not only if, but hardly if not) it appears later in the text in a newly developed context such as applications of graph theory via the progress of statements in the text. Thus, we assume a word that forms an *archipelago* makes an explanatory part of a document. Here, an archipelago is a sequence of *islands* of the occurrences of one certain word, where an island means the consecutive set of sentences where the word occurs frequently. We regard a word, of which the occurrences in the target text form an archipelago ranging across a length comparable to the target text, as a candidate of a keyword. This idea is based on the hypothesis that such a word is emphasized by the use in the island part (e.g., a paragraph) of the text, but it cannot be essential in the document if the word fails to appear again later. However, a stop word such as *is, a, are*, etc. (which are usually deleted first but here we aim to delete by the algorithm) are not considered as a keyword although they are distributed throughout the entire text, because they are just used to relate other words with each other. Stop words are often deleted in the preprocess of NLP including keyword extraction, but here let us challenge automated discounting of such a monotonously distributed word.

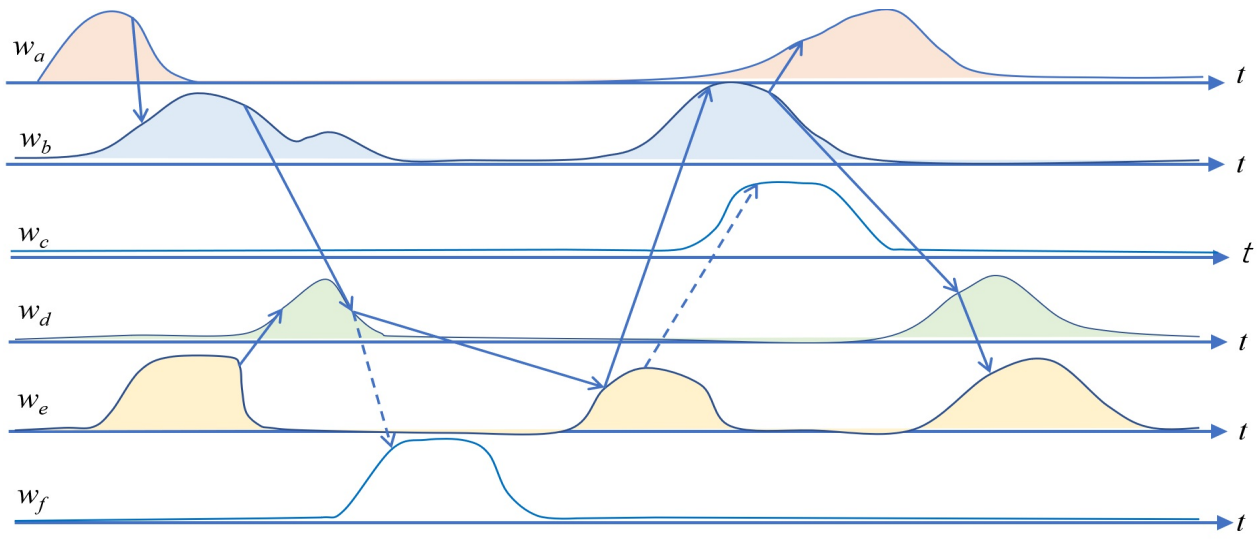

Figure 1: The archipelagoes of words call each other, causing the revival of explanatory words. The words forming single islands may be also called, but do not revive as these transient concepts play little roles in the explanatory structure of the text.

**Algorithm 1 Extraction of Word Archipelagoes**

**Input**: $D$ : the text with $n$ sentences where word $W$ occurs in the $s$-th sentence, $W$: the set of all words in $D$
**Parameter**: $\theta$, $\tau$
**Output**: $K$
1: Let $K = \phi$ (empty), $\Delta t = 1$.
2: **for all** $\Delta t < n$
3: **for all** $w \in W$
4: **while** $i\ \Delta t < n$ **do**
5: $t_i = \{s_{i\Delta t}, s_{i\Delta t+1}, \ldots, s_{(i+1)\Delta t}\}$
6: $f(w, t_i)$ = the counted number of $w \in t_i$
7: $p(w, i, \Delta t) = f(w, t_i)/\Sigma_i f(w, t_i)$
8: $H_A(w, \Delta t) = -\Sigma_i\, p(w, i, \Delta t) \log p(w, i, \Delta t)$
9: **if** $\Delta t > 2$ and $H_A(w, \Delta t) - H_A(w, \Delta t - 1) > H_A(w, 1)/n$ }
10: $\Delta t_{max_bound} = \Delta t$
11: **endif**
12: **if** $\Delta t_{max_bound} > \tau$ **then**
13: $K = K + \{w\}$
14: **endif**
15: **return** $K$.

An intuitive illustration of archipelagoes is in Figure 1. Each shadowed (colored in the colored version) sequence of islands shows an archipelago for one word having multiple islands, whereas the white ones ($w_c$ and $w_f$) are not archipelagoes. As depicted by solid arrows, an island in an archipelago for word $w_a$ calls for the occurrence of another concept shown by another word $w_b$ (indirectly via $w_c$ and $w_d$ in Figure 1), which calls back the occurrence of $w_a$ again if $w_a$ means a noteworthy concept for the entire document.

## 2.3 Window-based entropy

We propose the Algorithm 1 for extracting words forming archipelagoes. Below let us show the idea behind. Here we put the count of word $w$ in the $i$-th window by $f(w, t_i)$, and the distribution the count to all windows of width $\Delta t$ by

$$R(w, \Delta t) = [f(w, t_0), f(w, t_1), f(w, t_2), \ldots, f(w, t_n)]_{\Delta t} \qquad (1)$$

The window-based entropy $H_A$, or entropy A hereafter, of $w$ is defined in Eq.(2).

$$H_A(w, \Delta t) = -\Sigma_i\, p(w, i, \Delta t) \log p(w, i, \Delta t) \qquad (2)$$

where $p(w, i, \Delta t)$ represents $f(w, t_i)/\Sigma_i f(w, t_i)$, $\Sigma_i$ showing the sum for all $i$. A word $w$ forms an archipelago if Eq.(3) stands where $\theta$ and $\delta t$ (=1) represents a given positive value:

$$\exists \Delta t\ \{\ \Delta t > \tau,\ H_A(w, \Delta t) - H_A(w, \Delta t - \delta t) > \theta\ \} \qquad (3)$$

Eq.(3) means the value of $H_A$ decreases over a given threshold $\theta$ when the window is widened by $\Delta t$. We define this as the condition for word $w$'s organizing an archipelago, representing the following idea. That is, in Figure 2(a), the value of entropy A decreases until $\Delta t$ increases to a relatively small value e.g., from 1 to 2 here. On the other hand, in Figure 2(b), the value decreased until the increase in $\Delta t$ to a larger value e.g., from 4 to 8 here. Note $H_A$ stays constant for $\Delta t$ from 2 to 4 because each island is included in one window. In Figure 2(c), on the other hand, the single island is wide enough to cover a comparably large portion of the target text. Here, $H_A$ should decrease monotonously for the increase in $\Delta t$. Thus, in a sequence of multiple islands in an archipelago, a word sequence as in Figure 2 (b), the decrease in $H_A$ occurs with the increase in $\Delta t$ after the range where $H_A$ stays constant ($\Delta t$ from 2 to 4 in Figure 2(b)). Because the decrease in $H_A$ stops for the latter range, the decrease after that is expected to be larger than the average decrease. Thus, $\theta$ in Eq.(3) should be set comparable with but larger than $\max_{\Delta t}(H_A/n)$. As $H_A$ does not increase with $\Delta t$, this upper bound is set to $H_A(w, 1)/n$ that is obtained in the first cycle of Algorithm 1.

(a) A single island distribution $f(w, t)$, of word $w$

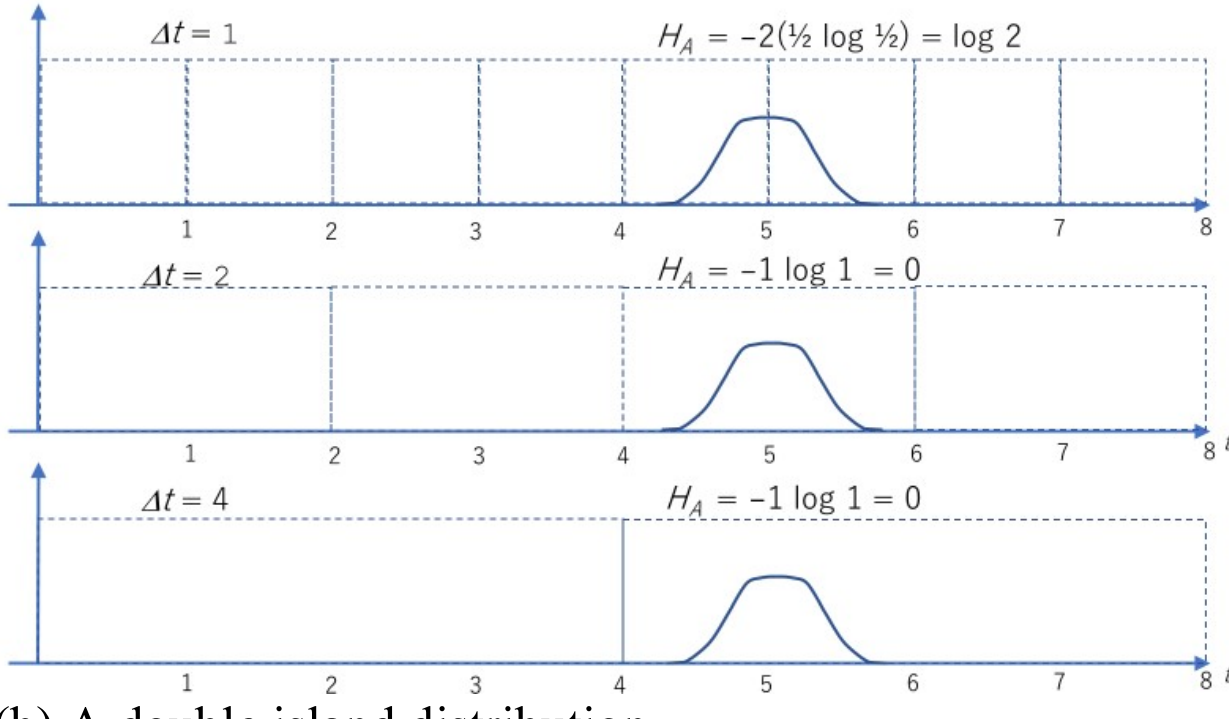

(b) A double island distribution

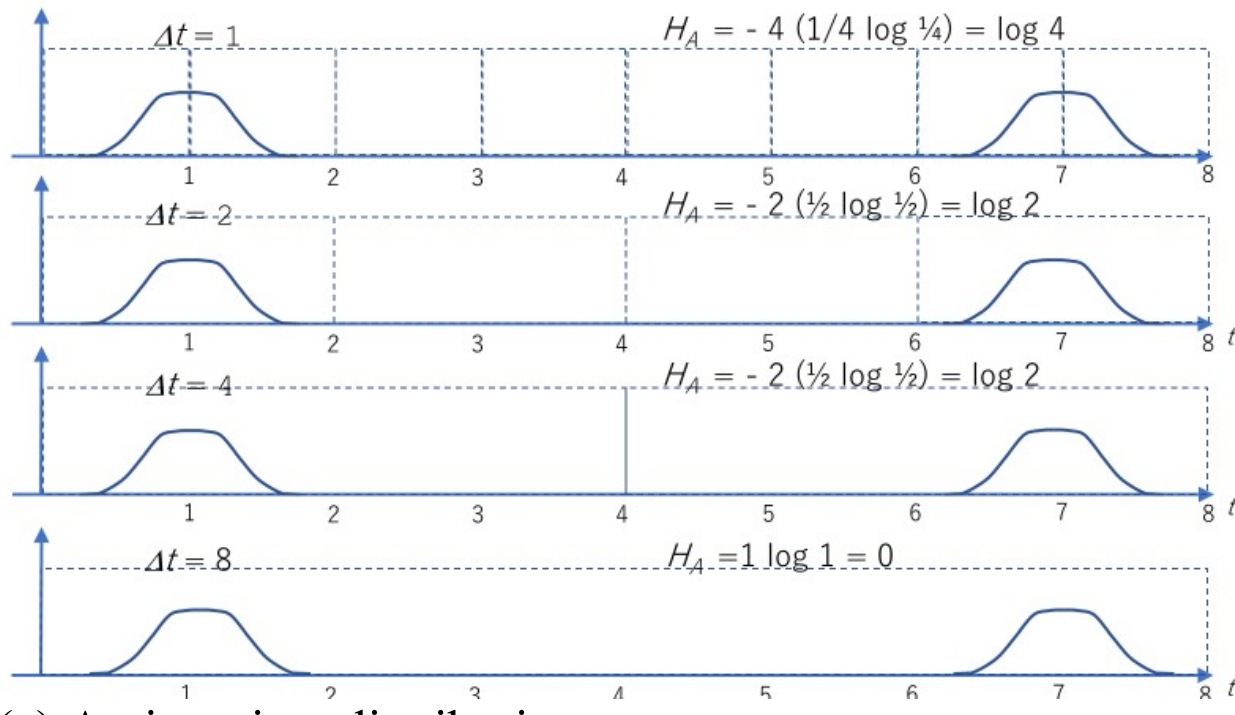

(c) An invariant distribution

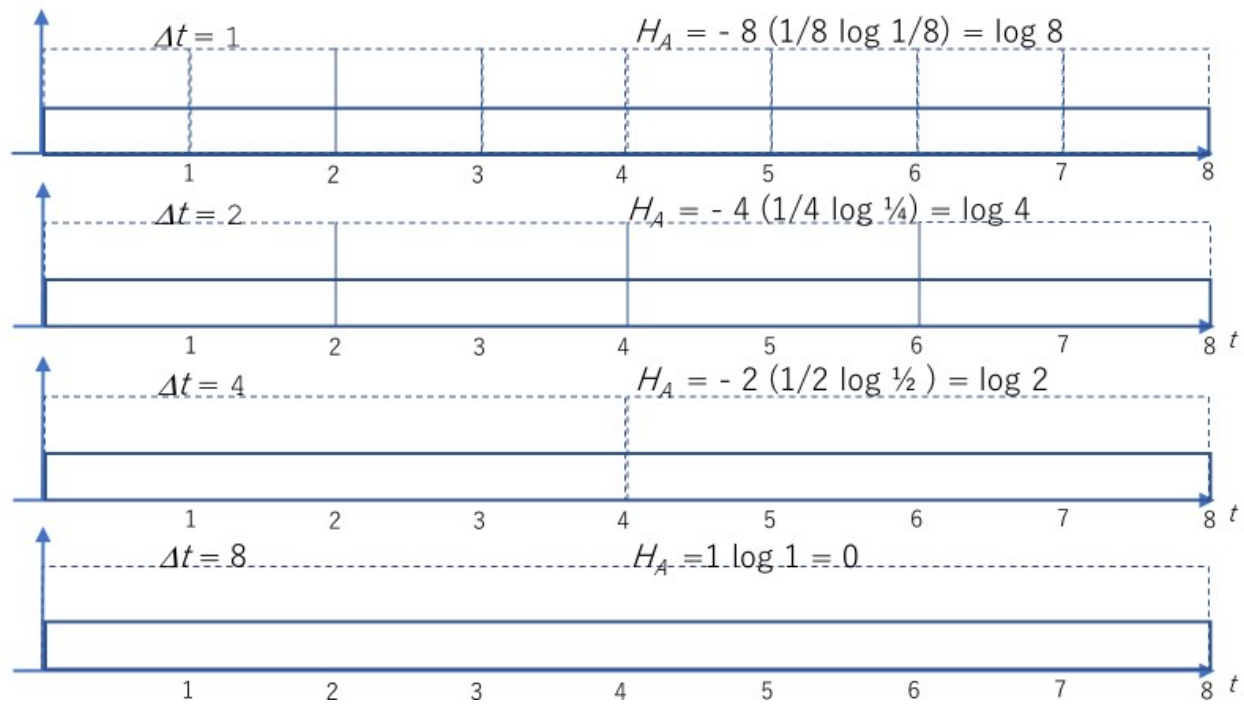

Figure 2: The distribution of words and the co-variation of entropy A with the window width

## 3 Evaluation of explanatory words

By visualizing the co-occurrences of words in the target text (put $D$' here), we can bridge each latent context represented by each cluster of items to other contexts. For the example above "*I have a book about mathematics. The concepts in discrete mathematics are hard to understand. One of the concepts is graph theory, ...*", the context of books appears first, and the topic of the entire domain of books is bridged to graph theory via ideas in mathematics. The sentences that follow may concentrate on graph theory. Thus, the clusters of words are connected via the explanatory flow in the text. Graph-based entropy (GBE), or entropy B defined as in Eq. (4), represents strong connections along the bridges among clusters to form a meta-cluster if the value is small.

$$H_B(D') = -\Sigma_c\, p(c) \log p(c) \tag{4}$$

Here $p(c)$ represents $f(c)/\Sigma_c f(c)$ where $f(w, t_i)$ represents the counted number of sentences in the $c$-th cluster, where a sentence is said to be in a cluster if it is closest to the cluster among other clusters. The measure of closeness here is defined by the cosine of two binary vectors, i.e., the word-based vector $s$: $(w_{s1}, w_{s2}, \cdots w_{sm})$ for sentence $s$ and $c$: $(w_{c1}, w_{c2}, \cdots w_{cm})$ for cluster $c$ . In $s_i$ and $c_i$, the presence of the $i$-th among $m$ words in $D$' is represented by 1 (the absence by 0). Then the value of entropy B for $D$' is computed on Algorithm 2 below. A cluster here means a group of items, connecting the top $\rho|W'|(|W'|+1)/2$ pairs of nodes via edges. The top pairs here mean pairs of the highest co-occurrence, where a co-occurrence shown as cooc($x$, $y$) is given by $p(x$ and $y)/\max(p(x), p(y))$ for a pair of items $x$ and $y$. This index of co-occurrence means the degree both $x$ and $y$ depend on each other's occurrence (differs from Symptom's coefficient, $p(x$ and $y)/\min(p(x), p(y))$, where the larger of the two dependency directions i.e. $x|y$ and $y|x$ is taken). Here, $p(x)$ is the proportion of sentences including $x$, $\rho$ is the given density of the co-occurrence graph to obtain, from which clusters above are taken as connected subgraphs. The less the value of $H_B$, the less separated the words are to different clusters. Thus, we use graph-based entropy $H_B$ as an index representing the logical explanatoriness of the extracted set of words that form archipelagoes on Algorithm A for the input text $D$'.

Note we are *not* using embedded vectors for two reasons here: first, one of our aims is to test if the proposed method stands alone, i.e., works without pre-learned knowledge or pattern related to each word's meaning. Second, the purpose is to make a method applicable to text including newly created contexts that may correspond to no existing concepts that can be learned from corpora. And, in obtaining clusters, we do not employ projection to the distance space of topics or distributed representation of words, because the aim of here is to represent the extent to which each word is logically explanatory, e.g., "*$w_1$ have $w_2$ about $w_3$. The $w_4$ in $w_2$ has a feature that $w_5$ is $w_6$. And the $w_5$ of $w_4$ is $w_6$ because...*" to be represented by a graph where $w_1$ and $w_3$ and $w_4$ are linked to $w_2$, and $w_4$ is in a cluster including $w_5$ and $w_6$. This explanation disappears if $w_4$ is deleted, which can be represented by the connection of nodes representing the words linked.

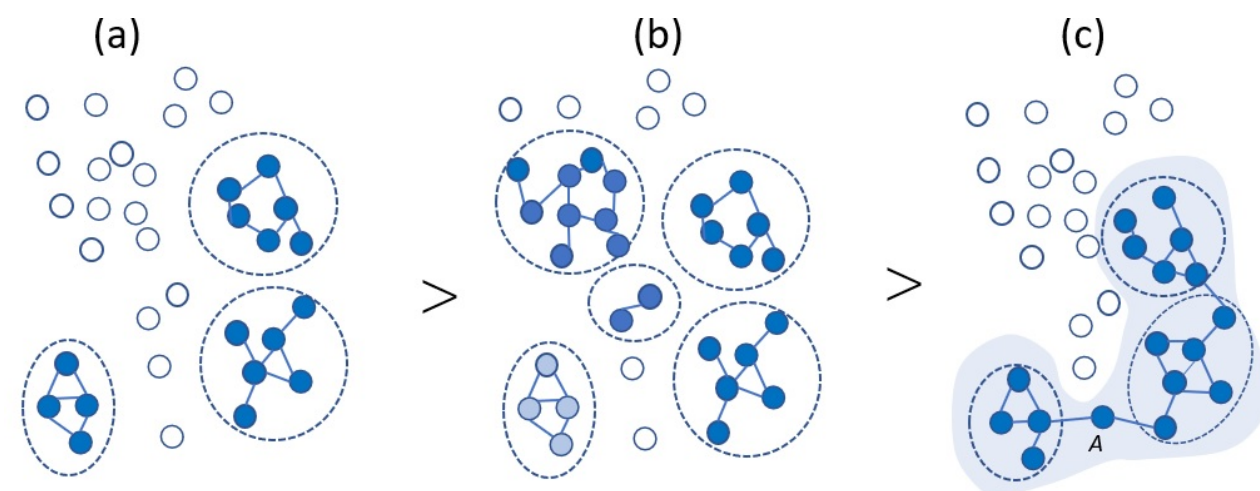

Figure 3: The value of $H_B$ is smaller if words are more connected, that is assumed to mean the text is explanatory. Word A in (c) substantially affects the logical structure of the text.

---

**Algorithm 2Extraction of Word Archipelagoes**

**Input**: *D'*: the text with $n$ sentences where word $W$ occurs in the $s$-th sentence, $W'$: the set of all words in $D$'
**Parameter**: $\rho$
**Output**: $H_B(D')$, the graph based entropy for $D$'
1: **for** $w_i \in W$
2: **for** $w_j \in W' | j > i$
3: **if rank.cooc**($w_i$, $w_j$) $\leq \rho|W'|(|W'|+1)/2$
4: cluster($w_j$) = cluster($w_i$)
5: **endfor**
**6:** $H_B(D') = -\Sigma_{c:\,\text{cluster}}\, p(c) \log p(c)$

---

In Figure 3, node A in (c) shows this kind of betweenness as of $w_4$ that tends to be pruned if the purpose is just clustering, but have been regarded to play an essential role in the context of forecasting or creating new trends in the target world [Ohsawa et al 2001, 2002, Ohsawa 2018], we choose to obtain clusters by connecting nodes via edges representing co-occurrence. Furthermore, we employ the hard clustering rather than soft for two reasons: first, because we should be sensitive to the cutting of a word such as node A in Figure 3 that means a loss of structure. Second, because the entropy in the strict sense can be obtained by assuming the clusters are exclusive i.e., not a word belongs to multiple clusters.

## 4 Results and Discussions

### 4.1 The procedure of experiments

First, each text $D$ has been processed with extracting the keywords using method $X$ ($X$ is chosen from the compared methods (1) to (5) below). The document is transformed into $D$' as a result, which has been evaluated using Algorithm B, by which the obtained entropy was compared. Here, Algorithm A is used as one of the compared methods.

We used the following texts created by famous authors or creative works rather than standardized sample documents used in text mining or NLP studies: William Shakespeare, Conan Doyle (the number of documents: 12), Moris Lebranc (32), Bertrand Russel (17), papers of IJCAI05 through 19 (40), and IEEE ICDM03 through 19 (10). We randomly chose the above number of papers from IJCAI and ICDM to make them comparable with other sets. This is because we focus on the extraction of explanatory words in creative documents that are supposed to induce new concepts where contextual shifts may be involved.

We extracted keywords by five methods **(1)** Algorithm 1 using $H_A$, setting the parameter values as $q$=0.2 and $\Delta t$ = 5, 10, 20 and baselines: **(2)** TFIDF using all the sample texts above for computing inverse document frequency (*idf*). **(3)** TextRank pruning the function to delete stop words because we evaluate the performance to ignore them automatically by the algorithm itself **(4)** RAKE algorithm **(5)** random choice of words in the text. For (2) through (5), the compared results were set to the same number of words extracted by (1).

## 4.2 The results of experiments

Before going to the results of the evaluation, let us see if the expected features of words have been found for explanatory words and words which are not. Figure 4 shows the distribution of each word in all the 133 sentences, for the exemplified text of Maurice LeBlanc's "Arrest of Lupin" in Rubber Gentleman Arsene Lupin (1905).

Based on the distribution in Figure 4, the variation of the window-based entropy for words is computed in Algorithm 1, of which the results are shown in Figure 5. According to the six subjects (who do not know any of the methods compared, aged 40 to 60) who read the text, words in Figure 5 (a) here are explanatory in this novel where Arsene Lupin has got arrested by Ganimard as a thief when he disguised himself as a gentleman d'Andrezy and was with lady Jerland in a boat on the ocean. The distribution of "Ganimard" and "d'Andrezy" in Figure 4 (a) and (b) are reflected in the discontinuous decrease in the value of window-based entropy (entropy A, $H_A$) in Figure 5 (a). On the other hand, the distribution of "the" in Figure 4 (c) is monotonous which is reflected in the continuous decrease in Figure 5 (b). Furthermore, the occurrence of "children" in Figure 4(d) is localized to a small part of the text, which is reflected in its staying at zero-value in Figure 5 (b). Neither of these patterns in Figure 5 (b) match with the condition to be archipelagoes (set $K$) in Algorithm A assumed to mean explanatory words.

**Evaluation: the comparison on entropy B.**

As shown in Table 1, the result of method (1) shows a constantly higher level than the randomly chose words i.e., the smaller value of $H_B$ for all the sample-sets. This tendency is strongest for $\tau$ set to 10, i.e., that corresponds to our daily intuition that important topics in documents appear before and after one or few paragraphs, and each paragraph includes 5 to 10 sentences. Thus, 20 is an over-distance for from one to the next frequently occurring part of the text for one word, and 5 is too small because 5 sentences tend to be included in one paragraph that can be regarded as one island in Figure 1.

Comparing (1) and (5: the random choice of words), we find (1) outperforms (5) significantly for most datasets. In the comparison shown in Table 1 with methods (2), (3), and (4), the number of extracted keywords has been set equal to the number for (1) setting $\Delta t$ to 10. This setting is chosen in this presentation because the performance as the percentage texts where $H_B$ was smaller in (1) than (5) was of the average among all $\Delta t$ (the last row of Table 1). As a result, (1) significantly outperformed (2: *tfidf*). This may be not an outstanding superiority of the proposed method because *tfidf* is a traditional method used for long, but the point is that method (1) developed here outperforms *tfidf* all in all, without any preparatory knowledge such as the corpus used in computing *tfidf* or the result of machine learning.

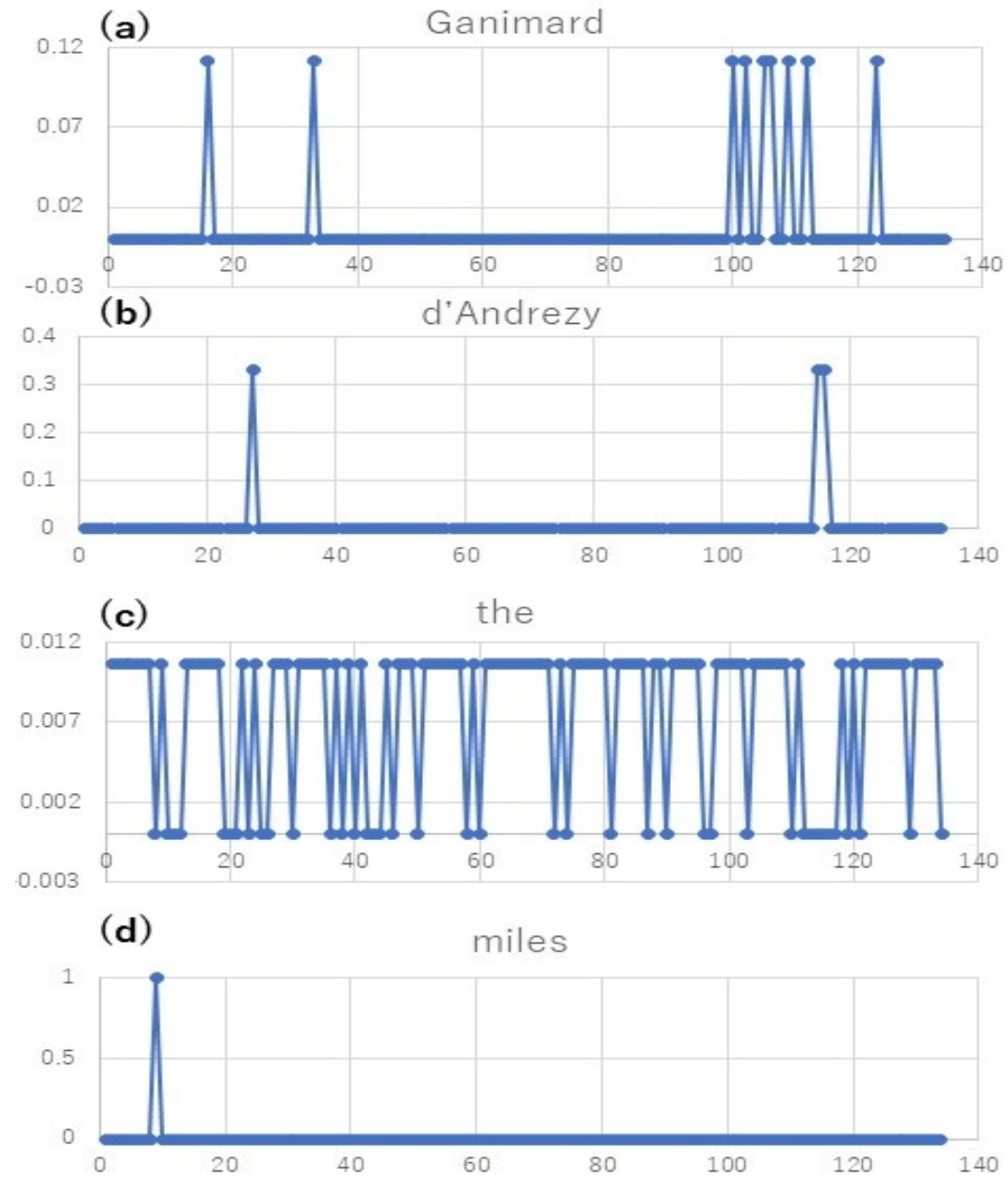

Figure 4: The distribution of words to sentences in "Arrest of Lupin" by Maurice LeBranc (1905): the $X$ axis shows the sentences in the order of appearance in the text.

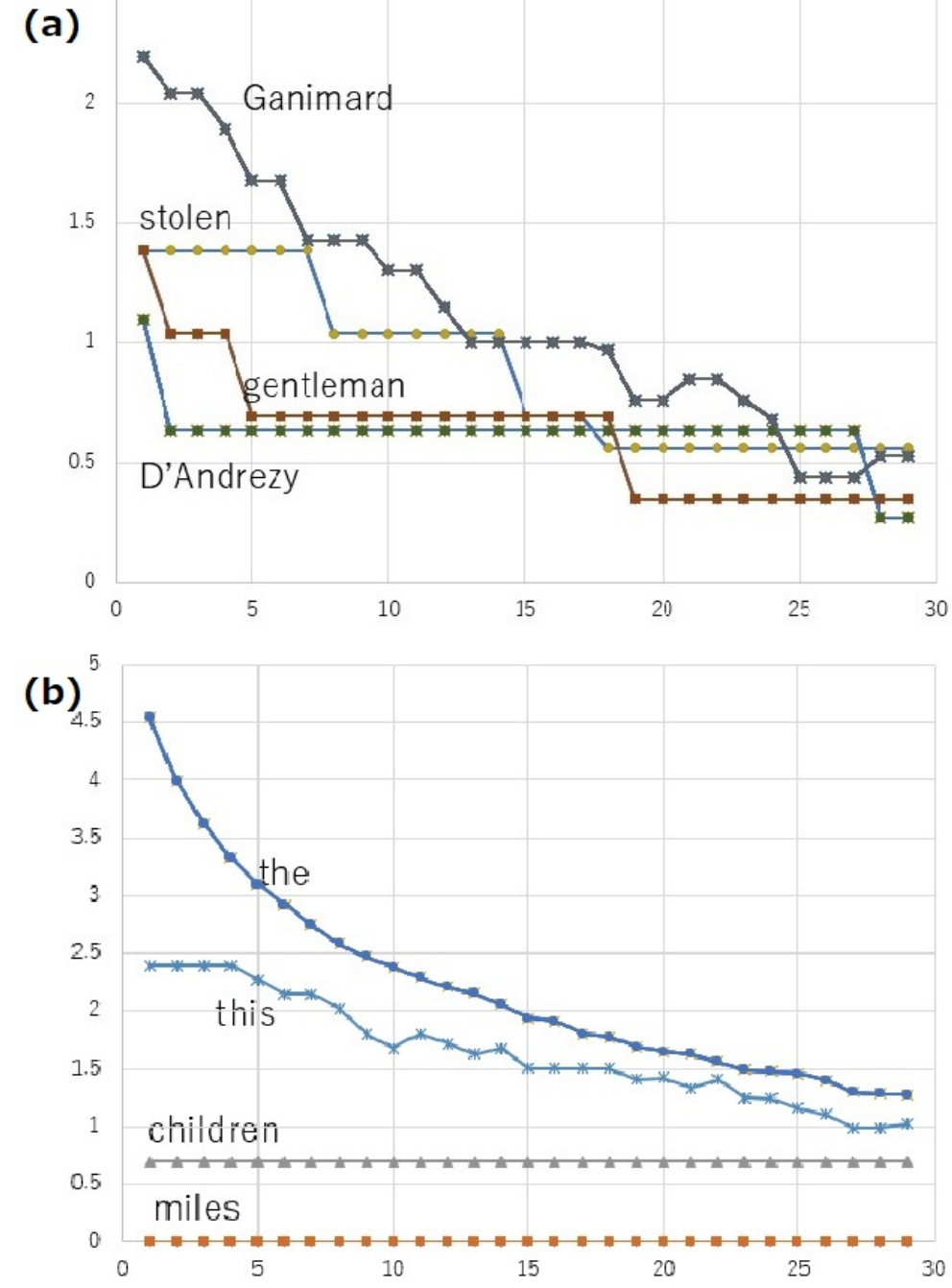

Figure 5: The variation of window-based entropy, X: the width $\Delta t$ of windows, Y: the value of entropy A. Here, (a) shows some words selected as in set $K$ in Algorithm 1, whereas (b) are not.

Also, (3: TextRank) tends to outperform *tfidf* mostly, interpreted to be because it models the interrelation of words that can be a result of the contextual flow structure of the text. However, this method tends to miss words which play the role as a bridge of influence, that is comprehensible because TextRank borrows the idea of PageRank where the targets rather than the bridges of interests are weighted and extracted. A similar tendency appears to (4: Rake) which models the relatively high degree nodes.

The reader may discuss such a word connecting islands to islands can be extracted as high-degree nodes or edges bridging clusters from the graph drawn in computing $H_B$. However, the importance of such a node $X$ as in an archipelago cannot be evaluated on the difference of the graph structure for $D$ with $X$ and $D'$ after cutting just $X$ away, because $X$ is tied to other words in islands and bridges in the archipelago of $X$. Therefore, the importance of $X$ is hardly evaluated without identifying the archipelago. This point can be intuitively understood from co-occurrence (cooc previously introduced) graph of extracted words as in Figure 6 where the 50 edges among 50 keywords are visualized for $\Delta t$ = 10. As shown here, the words extracted by Algorithm 1 such as "d'Andrezy" (dandrezy in the graph) and "Jerland" disappear, if they do, together with the connected words even if their frequencies are low (e.g., d'Andrezy which appeared only 3 times but is the disguised gentleman of Lupin).

| | (1) Algorithm 1 on $H_A$ for three values of $\Delta t$ below | | | (2) *tf idf* | (3) Text-Rank | (4) Rake |
|---|---|---|---|---|---|---|
| | **5** | **10** | **20** | **comp. with $\Delta t$ =10** | | |
| Doyle | 178 | **243** | 219 | 275 | 263 | 265 |
| (5) random | 309 | 301 | 277 | | | |
| LeBranc | 175 | **207** | 180 | 241 | 231 | 228 |
| (5) random | 306 | 267 | 210 | | | |
| Shake-speare | 178 | **239** | 237 | 246 | 242 | 250 |
| (5) random | 307 | 299 | 282 | | | |
| Grim | 175 | **208** | 158 | 223 | 212 | 217 |
| (5) random | 250 | 239 | 174 | | | |
| Andersen | 190 | 225 | 200 | **214** | 234 | 230 |
| (5) random | 312 | 240 | 225 | | | |
| Russel | 135 | 242 | 220 | 250 | **238** | 240 |
| (5) random | 283 | 275 | 231 | | | |
| IJCAI | 171 | **262** | 194 | 286 | 281 | 270 |
| (5) random | 272 | 320 | 202 | | | |
| ICDM | 172 | 243 | 191 | 258 | 240 | **220** |
| (5) random | 298 | 301 | 218 | | | |
| % of texts (1) < (5) | 98.5 | 92.1 | 75.5 | 84.4 | 85.0 | 87.6 |

Table 1: The comparison of $H_B$. The outperformance of $H_A$ is similar for the 3 settings of $\Delta t$. Each cell shows the average of $H_B$. The underlined and bold-letter values show smaller than (5) random on paired one-sided t-tests ($p<0.05$) and the largest values among the compared methods respectively.

## 5. Conclusions

Word archipelagoes and a method to extract them is here proposed. The islands meaning the local sub-sequence of sentences where a certain word is emphasized, and an archipelago connecting the islands, are regarded as parts of the text that represent the logical explanatory flow. Archipelagoes may look simple in comparison with previous methods where interactions among words are modeled explicitly using the structural features of the text such as edges in a graph, but are really the result of interactions among words as in Figure 1. We proposed to use window-based entropy for extracting word archipelagoes which interact with each other and organize the logical flow in the text. This expectation is here validated in comparison with other unsupervised methods for keyword extraction according to the evaluation using graph-based entropy (entropy B). Note, however, this method is not just a rival of the compared methods because each method can take its part of the explanatory flow in the text, that can be combined in real use.

This paper is just one step toward the explanation of the logical structure. Our future work shall include the detection of an important sequence of items in a dataset or words in a text without knowledge about the language or about the items. The connection of extracted sequence to create new knowledge is our long-term aim too.

(a) $H_A$ i.e., Algorithm 1

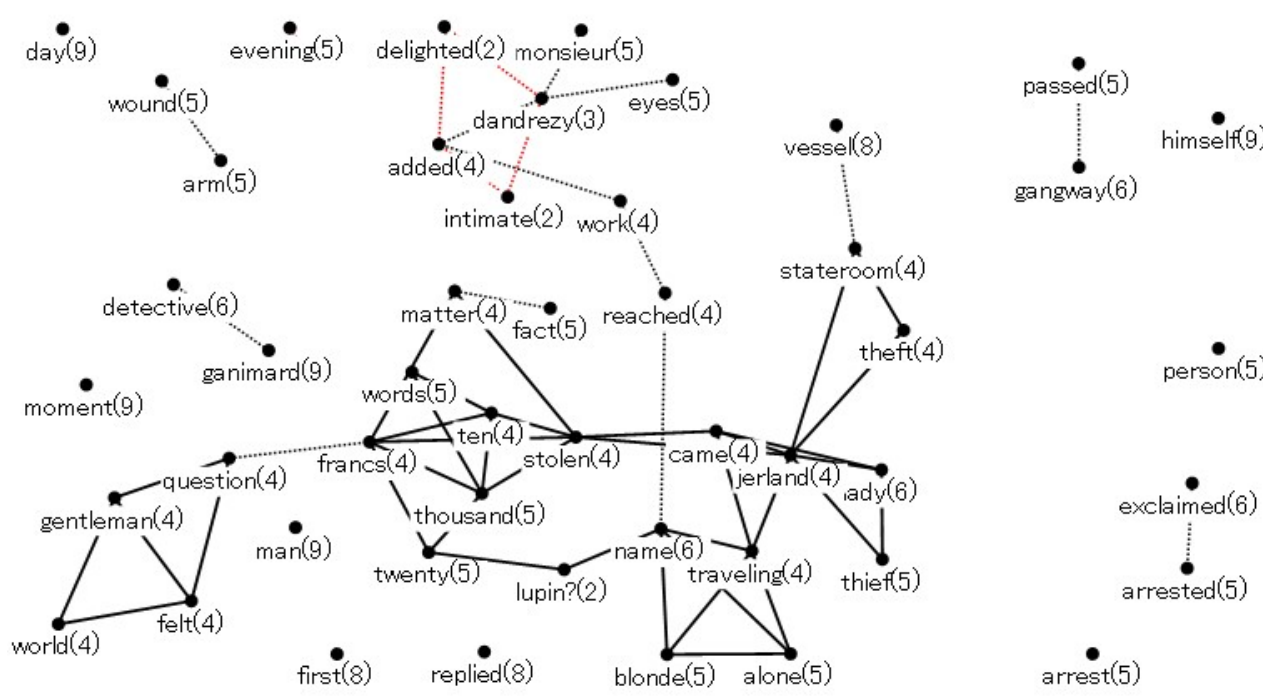

(b) *tfidf*

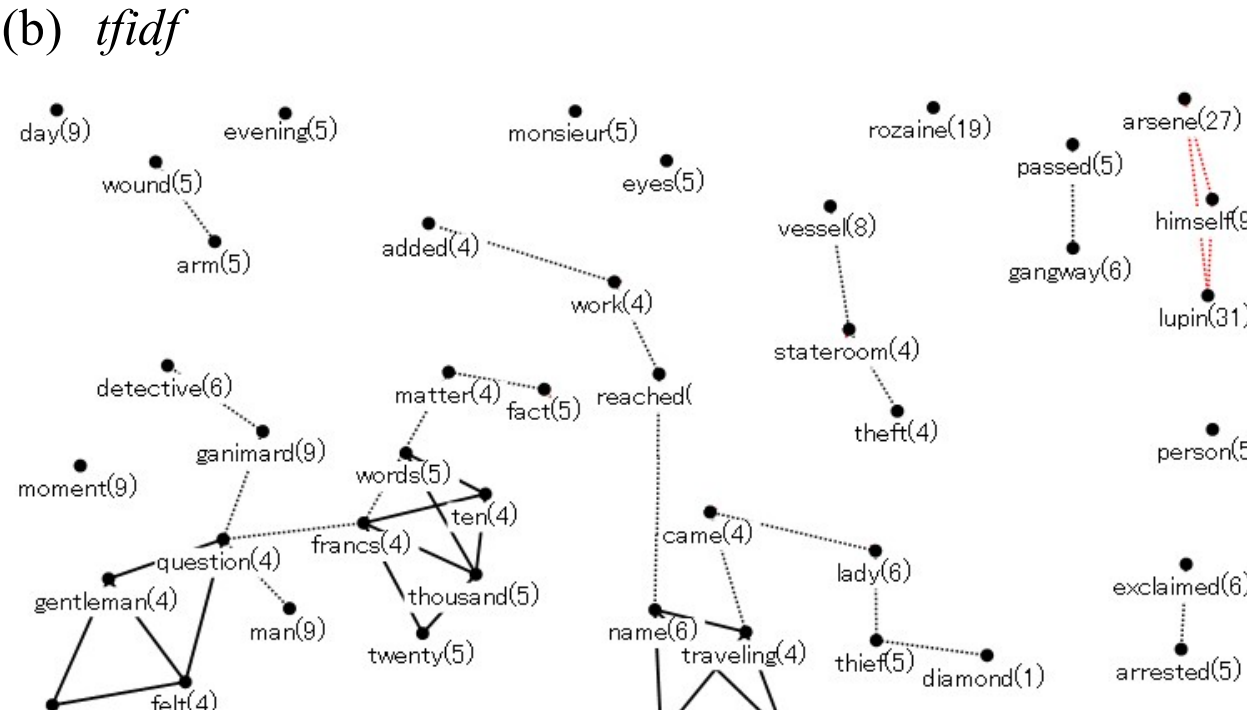

Figure 6: The co-occurrence of words extracted with (a) HA and (b) tfidf. Jerland and d'Andrezy that are of low frequency do not appear in (b), which results in hiding the structure (stop words are cut here).

## Acknowledgment

This study was supported by the Japan Society for the Promotion of Science Kakenhi (20K20482 and 23H00503).